\documentclass[10pt,twocolumn,letterpaper]{article}

\usepackage{iccv}
\usepackage{times}
\usepackage{epsfig}
\usepackage{graphicx}
\usepackage{amsmath}
\usepackage{amssymb}
\usepackage{subfiles}

\usepackage{multirow}
\usepackage{adjustbox}

\newcommand{\widthscalefive}{0.125}

\usepackage[final]{pdfpages}
\usepackage{booktabs}
\usepackage{subfiles}

\usepackage[pagebackref=true,breaklinks=true,letterpaper=true,colorlinks,bookmarks=false]{hyperref}

 \iccvfinalcopy 

\def\iccvPaperID{5397} 

\ificcvfinal\fi
\begin{document}

\title{ Hybrid Residual Attention Network for Single Image Super Resolution}

\author{Abdul Muqeet\\
Kyung Hee University\\
South Korea\\
{\tt\small amuqeet@khu.ac.kr}
\and
Md Tauhid Bin Iqbal\\
Kyung Hee University\\
South Korea\\
{\tt\small tauhidiq@khu.ac.kr}
\and
Sung-Ho Bae\\
Kyung Hee University\\
South Korea\\
{\tt\small shbae@khu.ac.kr}
}

\maketitle

\begin{abstract}
    The extraction and proper utilization of convolution neural network (CNN) features have a significant impact on the performance of image super-resolution (SR). Although CNN features contain both the spatial and channel information, current deep techniques on SR often suffer to maximize performance due to using either the spatial or channel information. Moreover, they integrate such information within a deep or wide network rather than exploiting all the available features, eventually resulting in high computational complexity. To address these issues, we present a binarized feature fusion (BFF) structure that utilizes the extracted features from residual groups (RG) in an effective way. Each residual group (RG) consists of multiple hybrid residual attention blocks (HRAB) that effectively integrates the multiscale feature extraction module and channel attention mechanism in a single block. Furthermore, we use dilated convolutions with different dilation factors to extract multiscale features. We also propose to adopt global, short and long skip connections and residual groups (RG) structure to ease the flow of information without losing important features details. In the paper, we call this overall network architecture as hybrid residual attention network (HRAN). In the experiment, we have observed the efficacy of our method against the state-of-the-art methods for both the quantitative and qualitative comparisons.

\end{abstract}

\section{Introduction}

In this paper, we address the Single Image Super Resolution (SISR) problem, where the objective is to reconstruct the accurate high-resolution (HR) image from a single low-resolution (LR) image. It is known as an ill-posed problem, since there are multiple solutions available for mapping any LR image to HR images. This problem is intensified when the up-sampling factor becomes larger. Because HR images preserve much richer information than LR images, SISR techniques are popular in many practical applications, such as surveillance \cite{zou2012very}, Face Hallucination \cite{yang2010image}, Hyperspectral imaging \cite{lanaras2015hyperspectral}, medical imaging \cite{shi2013cardiac} etc. 

Numerous deep learning based methods have been proposed in recent years to address the SISR problem. Among them, SRCNN \cite{dong2016image} is considered as the first attempt to come up with a deep-learning based solution with its three convolution layers. SRCNN outperformed the existing SISR approaches that typically used either multiple images with different scaling factors and/or handcrafted features. 
Later, Kim et al. \cite{dong2016accelerating} proposed an architecture named VDSR that extended the depth of CNN up to twenty layers while adding a global residual connection within the architecture. DRCN \cite{Kim_2016_DRCN} also increased the depth of network through a recursive supervision and skip connection, and improved the performance. 
However, due to increasing depth of the networks, vanishing gradient resisted the network to be converged \cite{he2016deep}. In the image classification domain, to solve the aforementioned problem, He et al. \cite{he2016deep} proposed a residual block by which a network over 1000 layers was successfully trained. Inspired by its very deep architecture with residual blocks, EDSR \cite{lim2017enhanced} proposed much wider and deeper networks for the SISR problem using residual blocks, called EDSR and MDSR \cite{lim2017enhanced}, respectively. 

Very recently, Zhang et al. \cite{zhang2018image} proposed RCAN that utilizes a channel attention block to exploit the inter-dependencies across the feature channels. Moreover, Li et al. \cite{Li_2018_ECCV} proposed MSRN that improved the reconstruction performance by exploiting the information of spatial features rather than increasing the depth of CNNs. MSRN combines the features extracted from different convolution filter sizes and concatenates the outputs of all residual blocks through a hierarchical feature fusion (HFF) technique, utilizing the information of the intermediate feature maps. By doing so, MSRN achieved comparable performance against EDSR \cite{lim2017enhanced} although having  a 7-times smaller model size. In \cite{zhang2019dcsr}, Zhang et al. proposed DCSR in which they proposed a mixed convolution block that combines dilated convolution layers and conventional convolution layers to attain larger receptive field sizes.
Nonetheless, most of these CNN-based methods focused either on increasing the number of layers~\cite{kim2016accurate, Kim_2016_DRCN, lim2017enhanced, zhang2018image} or on extending the width and height in a layer of CNN to achieve higher performance~\cite{Li_2018_ECCV}. In this way, they put less focus on exploiting the by-product CNN features, \eg, spatial and channel information, simultaneously, and thus suffer to maximize the performance at times.

Moreover, the strong correlations between the input LR and output HR images~\cite{Li_2018_ECCV} lead us to making an assumption that, apart from the high-level features, the both low-level and mid-level features also play vital roles  for reconstructing an super-resolution (SR) image. Therefore, we argue that, they should be treated precisely in this paper. 

In the previous work, dense connections were used ~\cite{tong2017image}, which added every feature to subsequent features with residual connections. 
As a variant of dense connections, hybrid feature fusion (HFF) ~\cite{tong2017image, zhang2018residual, Li_2018_ECCV} was proposed to remove the trivial residual connections and to directly concatenate all the output features from the residual blocks for the SISR problem. However, this direct feature concatenation prohibit the features from smooth feature transformation from low to high levels, resulting in resulting in improper utilization of various low-level and mid-level features. This may  introduce redundancy in feature utilization, thus increasing the cost of computation complexity. In our ablation study in Section 4.1, this problem will be verified.

To solve this problem, in this paper, we propose a binarized feature fusion (BFF) scheme that combines adjacent feature maps with $1\times1$ convolutions which are repeatedly performed until remaining a single feature map. This allows all the features extracted from the CNN to be integrated smoothly, thus fully utilizing various features with different levels. Moreover, to efficiently extract the features, unlike previous work that used main residual blocks, we propose to use residual groups (RG) that constructs with the proposed hybrid residual attention block (HRAB). 
Our proposed HRAB extracts both the spatial and channel information with the notion that the both information is important in the reconstruction of high quality SR images and should be extracted simultaneously in a single module.

Moreover, compared to MSRN ~\cite{Li_2018_ECCV} that concatenates the conventional convolution layers with different kernel sizes to enlarge receptive field sizes, proposed method concatenates dilated convolution layers with different dilation factors exploiting much larger receptive fields while significantly decreasing the number of parameters, \ie, convolution weights. Furthermore, to ease the flow of information, we introduce the short, long and global skip connections. We conduct comprehensive experiments to verify the efficacy of our method, where we observe its superiority against other state-of-the-art methods.

We summarize the overall contributions of this work as, 
\begin{itemize}
  \item We propose a BFF to transfer all the images features smoothly by the end of the network. This structure allows the network to smoothly transform the features with different levels and generate an effective feature map in the final reconstruction stage. 
  \item We propose a hybrid residual attention block (HRAB) that considers both channel and spatial attention mechanisms to exploit the channel and spatial dependencies. The spatial attention mechanism extracts the fine spatial features with larger receptive field sizes whereas the channel attention guides in selecting the most important feature channels thus in the end, we have more discriminative features.
  \item Other than previous works, we employ BFF on residual groups (RG) rather than residual blocks (HRAB)
  \item For extracting the multiscale spatial features, we propose to use a mixed dilated convolution block with different dilation factors. Compared to the previous work in \cite{Li_2018_ECCV} that used the large kernel sizes to secure large receptive fields, our proposed method can achieve a similar performance even with smaller kernel sizes. Moreover, we propose to use the dilated convolution in an effective manner to avoid the gridding problem of the conventional dilated convolution layers.
  \item To ease the transmission of information through out the network, we propose to adopt the global, short and long skip in our architecture.
\end{itemize}


\section{Related work}

There are several CNN-based SISR methods that have been proposed in the recent past. Previously, in the preprocessing step, researchers tend to use an interpolated LR image as an input that is interpolated to desired output image size which enables the network to have the same size of input and output images. In contrast, due to the additional computation complexity of interpolation, current work emphasizes to directly reconstruct HR image from LR image without interpolation. 

In 2014, Dong et al. \cite{dong2016image} proposed SRCNN, the first CNN network architecture in the SR domain. It was a shallow 3 layers CNN architecture which achieved the superior performance against the previous non-CNN methods. Later, He et al. \cite{he2016deep} proposed a residual learning technique, and then Kim et al. \cite{kim2016accurate, Kim_2016_DRCN} achieved remarkable performance with their proposed VDSR and DRCN methods. VDSR used the deep (20 layers) CNN and global residual connection whereas DRCN \cite{Kim_2016_DRCN} used a recursive block to increase the depth that does not require new parameters for repetitive blocks. Tai et al. \cite{tai2017memnet} proposed the MemNet which had memory blocks that consist of recursive and gate units. All of these methods have used the interpolated LR image as input. Due to this preprocessing, these methods add additional computation complexity along with artifacts, as also described in \cite{shi2016real}. 


\begin{figure*}
\begin{center}
\includegraphics[scale = 0.18]{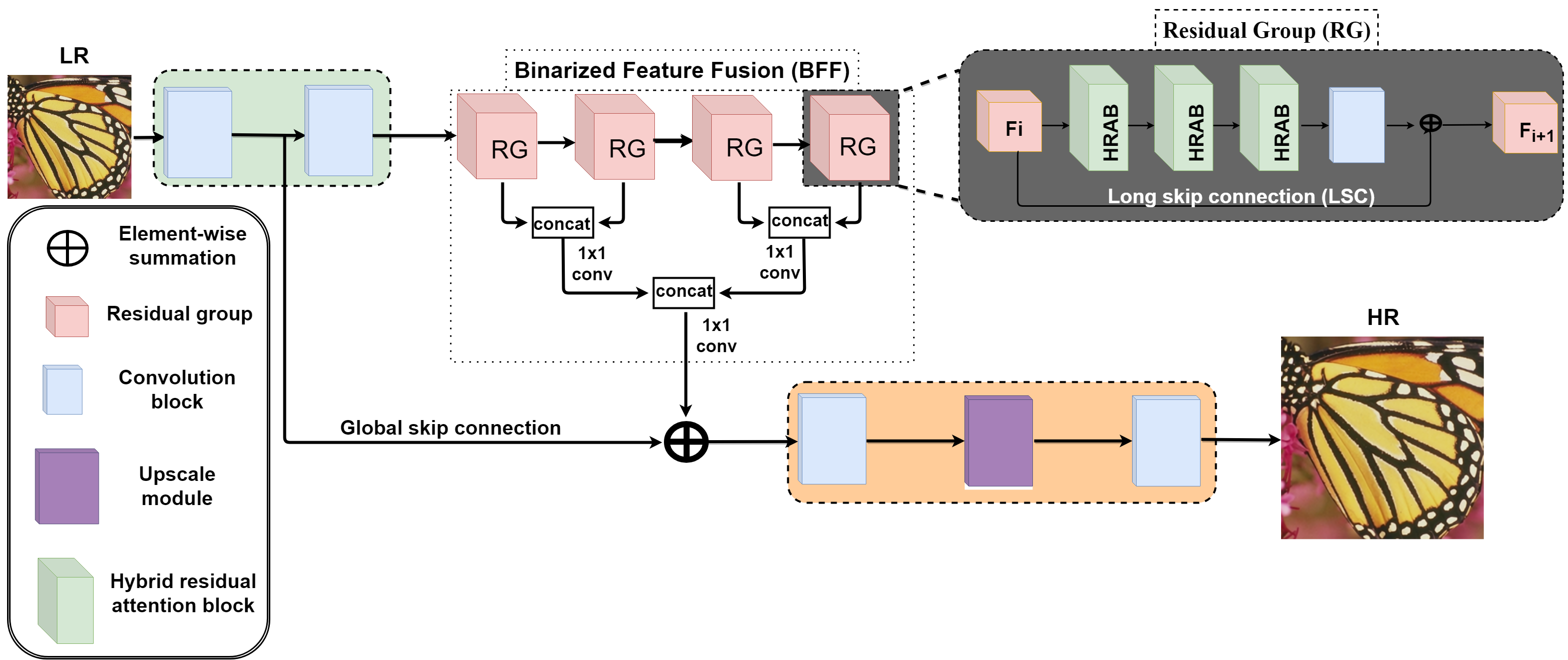}
\end{center}
   \caption{
   The proposed network architecture HRAN. The green-shaded area at top-left performs shallow feature extraction. the gray-shaded area at top-right indicates the internal structure of RG. The proposed BFF smoothly integrates features from low to high level RG blocks, and the output of BFF is element-wise summed with the shallow features and is fed into the final reconstruction stage (the orange-shaded area) to produce an HR image. The left-bottom block shows the specific descriptions.}
\label{fig:HRAN}
\end{figure*}

On the other end, the recent state-of-the-art methods directly learn the mapping from input LR image. Dong et al. \cite{dong2016accelerating} proposed the FSRCNN, an improved version of SRCNN, having faster training and inference time. Ledig et al. \cite{ledig2017photo} proposed the SRResNet, inspired from ResNet \cite{he2016deep}, to construct the deeper network. With the perceptual loss function in GAN, they proposed the SRGAN for photo-realistic SR. Lim et al. \cite{lim2017enhanced} removed the trivial modules (like batch normalization) of SRResNet, and proposed the EDSR (wider) and MDSR (deeper) that made a significant improvement in SR problem. EDSR has a large number of filters (256) whereas MDSR has a small number of filters though the depth of CNN network is increased to about 165 layers. It also won the first NTIRE SR challenge \cite{timofte2017ntire}. It has shown that deeper networks can achieve remarkable performance. Consequently, Zhang et al. \cite{zhang2018image} proposed a very deep network for SR. To the extent of our knowledge, it has the largest depth in the SR domain. RCAN \cite{zhang2018image} has shown that only stacking the layers cannot improve the performance. It proposed to use the channel attention (CA) \cite{hu2018squeeze} mechanism to neglect the low-frequency information while selecting the valuable high-frequency feature maps. To increase the depth of the network, it proposed the residual in residual (RIR) structure. Nevertheless, RCAN \cite{zhang2018image} network is very deep and makes it difficult to use it in real-life applications due to higher inference time. 

In contrast, multiscale feature extraction technique, which is less explored in SISR, has shown significant performance in object detection, \cite{lin2017feature} image segmentation, \cite{seferbekov2018feature} and model compression \cite {chen2018big} to achieve good tradeoffs between speed and accuracy. Li et al. proposed a multiscale residual network (MSRN) ~\cite{Li_2018_ECCV} having just 8 residual blocks. It used multipath convolution layers with different kernel sizes (3$\times$3 and 5$\times$5) to extract the multiscale spatial features. Furthermore, it proposed to use the hierarchical feature fusion (HFF) architecture to utilize the intermediate features. The intuition behind HFF architecture is to transfer the middle features at the end of the network since an increase in the depth of the network may cause the features to vanish in between the network. HFF shows comparable performance to EDSR, nevertheless its accuracy is limited. In addition, as the depth or width of a network increases, HFF also increases the computation complexity.

Therefore, we need an efficient multiscale superresolution CNN which could fully utilize the feature information as well as channel information. Considering it, we propose a hybrid residual attention network (HRAN) which combines the multiscale feature extraction along with the channel attention \cite{hu2018squeeze} mechanism. In this paper, we refer the multiscale feature extraction as spatial attention. Thus, the combination of the channel and spatial attention is called hybrid attention. We discuss the details of HRAN in the next section.

\section{Hybrid Residual Attention Network}

\subsection{Network architecture} 

The proposed HRAN architecture is shown in Figure~\ref{fig:HRAN}. The HRAN can be decomposed into two parts: feature extraction and reconstruction. The feature extraction is divided into two parts: shallow feature extraction and deep feature extraction. The deep feature extraction step further includes residual groups (RG) with binarized feature fusion (BFF) structure. Whereas, RG contains a sequence of hybrid residual attention blocks (HRAB) followed by 3$\times$3 convolution. We represent the input and output of HRAN as $I_{LR}$ and $I_{SR}$ respectively. We aim to reconstruct the accurate HR image $I_{HR}$ directly from LR image $I_{LR}$. 

In the shallow feature extraction, we use two convolution layers to extract the features from input $I_{LR}$ image.
\begin{align}
\begin{split}
\label{eq:SFE1}
F_{0}=H_{SF1}\left ( I_{LR} \right ),
\end{split}
\end{align}
Here $H_{SF1}\left ( \cdot  \right )$ represents the convolution operation. $F_{0}$ is also used for global residual learning to preserve the input features. As mentioned above, we pass the $F_{0}$ for further feature extraction
\begin{align}
\begin{split}
\label{eq:SFE2}
F_{1}=H_{SF2}\left ( F_{0} \right ),
\end{split}
\end{align}
\noindent where $H_{SF2}\left ( \cdot  \right )$ represents the convolution operation. $F_{1}$ is the output of shallow feature extraction step and will be used as input for the deep feature extraction. 
\begin{align}
\begin{split}
\label{eq:DF}
F_{DF}=H_{DF}\left ( F_{1} \right ) + F_{0} ,
\end{split}
\end{align}
Here $H_{DF}\left ( \cdot  \right )$ represents the deep feature extraction function and $F_{0}$ shows global residual connection like VDSR \cite{kim2016accurate} at the end of deep features. The deep features are sequentially extracted through HRAB, RG and BFF. The details are mentioned in later sections.
\begin{align}
\begin{split}
\label{eq:REC}
I_{SR}=H_{REC}\left ( F_{DF} \right ),
\end{split}
\end{align}
\noindent where $H_{REC}$ denotes the reconstruction function. For the image reconstruction, previously researchers upsampled the input image to get the desired output dimensions. we reconstruct the $I_{SR}$ having similar dimensions as $I_{HR}$ through deep features of $I_{LR}$. There are various techniques to serve as upsampling modules, such as PixelShuffle layer \cite{shi2016real}, deconvolution layer \cite{dong2016accelerating}, nearest-neighbor upsampling convolution \cite{dumoulin2017learned}. In this work, we use the MSRN ~\cite{Li_2018_ECCV} reconstruction module that enables us to upscale to any upscale factor with minor changes.

We can write the proposed HRAN function as 
\begin{align}
\begin{split}
\label{eq:HRAN}
I_{SR}=H_{HRAN}\left ( I_{LR} \right ),
\end{split}
\end{align}

For the optimization, numerous loss functions have been discussed for SISR. The mostly used loss functions are the MSE, L1, and L2 functions whereas perceptual and adversarial losses are also preferred. To keep the network simple and avoid the trivial training tricks, we prefer to optimize with L1 loss function. Hence, we can define the objective function of HRAN as :
\begin{align}
\begin{split}
L\left ( \Theta  \right )=\frac{1}{N}\sum_{i=1}^{N}\left \| H_{HRAN}\left ( I_{LR}^{ i } \right )-I_{HR}^{ i } \right \|_{1},
\end{split}
\end{align}
\noindent where $\Theta$ denotes the weights and bias of our network.



\begin{figure*}
\begin{center}
\includegraphics[scale = 0.15]{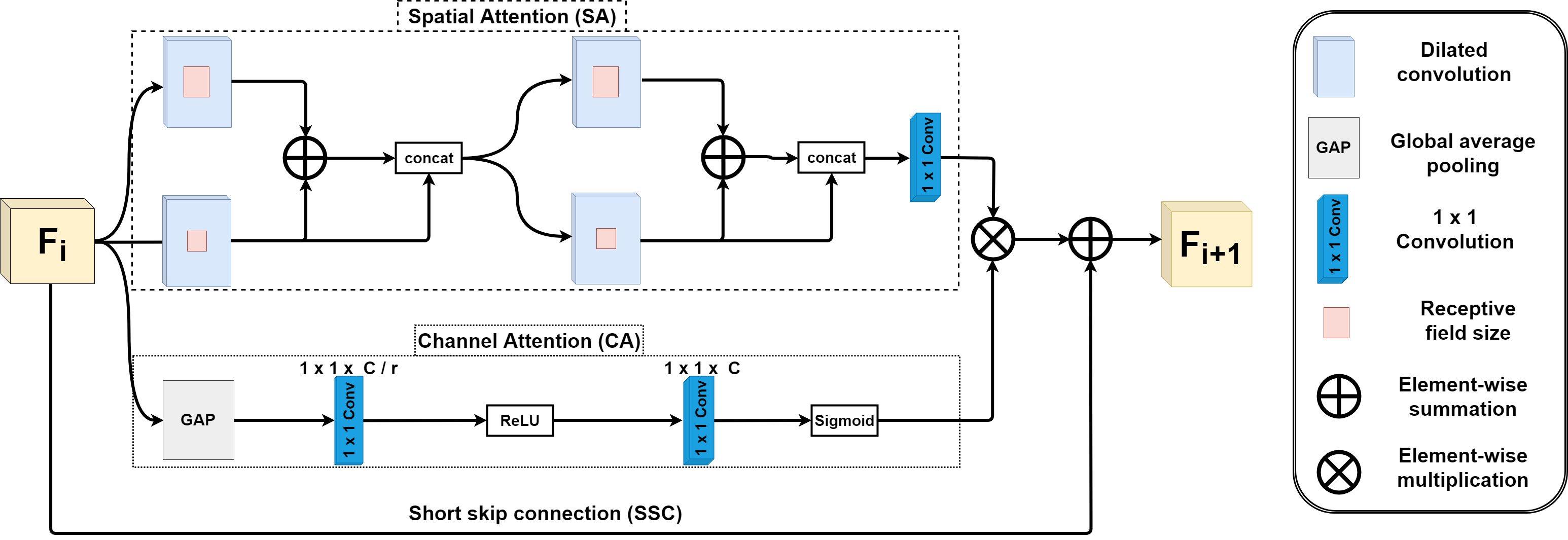}
\end{center}
   \caption{Proposed multi-path hybrid residual attention block (HRAB). Top path represents Spatial Attention (SA) that contains dilated convolutions with different dilation factors. Bottom path represents Channel Attention (CA) mechanism. Notations about different components are given in the right.}
\label{fig:HRAB}
\end{figure*}

\subsection{Binarized Feature Fusion (BFF) structure}

The shallow features lack the fine details for SISR. We use deep networks to detect such features. However, in SISR, there is a strong correlation between $I_{LR}$ and $I_{SR}$. It is required to fully utilize the features of $I_{LR}$ and transmit them to the end of network, but due to the deep network, features start gradually vanishing during the transmission. The possible solution is to use a residual connection, however, it induces the redundant information ~\cite{Li_2018_ECCV}. MSRN ~\cite{Li_2018_ECCV} uses the hierarchical feature fusion structure (HFF) to transmit the information from all the feature maps towards the end of the network. The concatenation of every feature generates a lot of redundant information and also increase the memory computation.

In contrast, we propose a binarized feature fusion (BFF) structure that as shown in Figure ~\ref{fig:HRAN}. The notable difference in this architecture is the use of residual groups (RG) instead of Multiscale residual block (MSRB)\cite{Li_2018_ECCV}. It is called a residual group due to residual connections within itself \ie, RGs are connected through LSC whereas its sub-module HRABs are connected through SSC. The use of RGs does not only help to increase the depth but also reduce the memory overhead when concatenating the features map. 

Another difference in this architecture is the feature extraction from adjacent RG blocks. First, we concatenate the adjacent RG blocks and then, we remove the redundant information from adjacent blocks using 1$\times$1 convolution. We repeat this procedure for all RG blocks and the resultant blocks produced through this mechanism until all the blocks integrate into single RG block, which is convolved by 1$\times$1 to produce the output features. In the end, we element-wise add this output to the shallow features' output ($F_{0}$). We refer this element-wise summation as global skip connection in Figure \ref{fig:HRAN}.

\begin{align}
\begin{split}
\label{eq:RIR}
F_{i+1}=H_{RG}\left ( F_{i} \right ),
\end{split}
\end{align}
 \noindent where $H_{RG}\left ( \cdot  \right )$ represents the features extracted through single RG block whereas $F_{i}$ shows the $i^{th}$ extracted feature map. We explain the details of RG in the next section. When we extract all the features through RG blocks, then we can utilize these RG blocks with HFF architecture.

\begin{align}
\begin{split}
\label{eq:H11_1}
M_{j}=H_{1\times1}\left  [ F_{i+1},F_{i+2}\right ] ,
\end{split}
\end{align}
\begin{align}
\begin{split}
\label{eq:H11_2}
M_{j+1}=H_{1\times1}\left  [ F_{i+3},F_{i+4}\right ] ,
\end{split}
\end{align}

Here, the output of two adjacent RG blocks are channel-wise concatenated and then passed into 1x1 convolution layer to avoid the redundant information from them. Thus, the four RG blocks produce two more blocks which are then processed in a similar manner such that $F_{i+1}=M_{j}$ and $F_{i+2}=M_{j+1}$. Thus, in the next step, $M_{j}$ and $M_{j+1}$ will act as two RG blocks. We repeat this procedure until we integrate all the RGs and resultant blocks into a single output which is further used in the input of reconstruction step.

\subsection{Residual Groups (RG)}

It is shown in \cite{lim2017enhanced} that the stacked residual blocks enhance the performances of SR but after some extent, cause crucial information loss during transmission and also makes the training slower, affecting the performance gain in the SISR \cite{zhang2018image}. Thus, rather than increasing the depth, we propose to use the residual groups (RG) (see shaded area of Figure ~\ref{fig:HRAN}) in our architecture to detect deep features. The RG consists of multiple HRAB that are followed by 1$\times$1 convolution. We find that adding many HRAB does degrade the SR performance. Thus, to preserve the information, we apply element-wise summation between the input of RG and output of 1$\times$1 convolution and refer it as long skip connection (LSC). 

The RG enables the network to remember the information through LSC whereas to detect deep features, it uses SSC within its modules, in this case, HRAB. Hence, the flow of information in RG is smoothly carried out through LSC and SSC. The details of the HRAB are mentioned in the next section. 

Thus, we express the single RG block as 
\begin{align}
\begin{split}
\label{eq:RIR2}
H_{RG}= W_{RG} * H_{n}\left (H_{n-1} \left  ( \cdots H_{1}\left  ( F_{1}\right )\cdots \right )\right ),
\end{split}
\end{align}

Here $H_{i}$ represents the `B' hybrid residual attention blocks (HRAB), which takes input features from previous RG block ($F_{i}$) and produces the output ($F_{i+1}$). After stacking the `B' HRAB modules, we apply 3$\times$3 convolutions with weights $W_{RG}$. After applying LSC, the equation~\ref{eq:RIR2} can be rewritten as 
\begin{align}
\begin{split}
\label{eq:RIR2_1}
H_{RG}= W_{RG} * H_{n}\left (H_{n-1} \left  ( \cdots H_{1}\left  ( F_{1}\right )\cdots \right )\right ) +  F_{1},
\end{split}
\end{align}
The above equation represents the first RG block because it takes the shallow features $F_{1}$ as input. Since, we have multiple RG blocks to extract the deep features, hence, the above equation can be generally written as
\begin{align}
\begin{split}
\label{eq:RIR2_2}
H^{i}_{RG}= W^{i}_{RG} * H^{i-1}_{RG} + H^{i-1}_{RG}
\end{split}
\end{align}
Here $i=1,2,\cdots,R.$ We have `R' RG blocks and each RG block uses the output of the previous block as its input except the first RG block that uses the shallow features $F_{1}$ as input. Thus, for the first RG block, $H^{0}_{RG}$ = $F_{1}$.




\subsection{Hybrid Residual Attention Block (HRAB)}

In this section, we propose a multiscale multipath residual attention block for the feature extraction, called hybrid residual attention block (HRAB) (see Figure ~\ref{fig:HRAB}). Our HRAB integrates both the spatial attention (SA) and channel attention (CA) mechanisms, thus, it has two separate paths for the SA and CA.
\vspace{-2mm}
\begin{align}
\begin{split}
\label{eq:HRAB}
H_{HRAB}\left  (F_{i+1}\right ) = H_{SA} \left  (  F_{i} \right ) \cdot H_{CA} \left  ( F_{i}\right )  
\end{split}
\end{align}

\noindent where $H_{SA}$ and $H_{CA}$ denote the functions of spatial attention (SA) and channel attention (CA) respectively. Here `$\cdot$' represents the element-wise multiplication between SA and CA functions. Unlike RCAN \cite{zhang2018image}, we propose to use element-wise multiplication between the outputs of SA and CA to extract the most informative spatial features. Like RCAN \cite{zhang2018image}, we also add the short skip connections (SSC) to ease the flow of information through the network.
\vspace{-3mm}
\subsubsection{Spatial Attention (SA)}

MSRN~\cite{Li_2018_ECCV} proves that multiscale features improve the performance with lesser residual blocks. In MSRN ~\cite{Li_2018_ECCV}, authors use the multiple CNN filters with increasing kernel sizes ($3\times3$ and $5\times5$) to extract multiscale features. The intuition behind the larger kernel size is to take advantage of large receptive fields. But, the large kernel size causes to increase the memory computation. Thus, we propose to use the dilated convolution layers with different dilation factors which can have the same receptive fields as large kernel size and memory consumption is similar to smaller kernel size. But, only stacking the dilated convolution layers produce gridding effect \cite{yu2017dilated}. To avoid this problem, as illustrated in Figure~\ref{fig:HRAB}, we propose to use the element-wise sum operation between the dilated convolutions with different factors before the concatenation operation. Suppose $F_{i-1}$ is the input of SA then the output will be $F_{i}$.

\begin{gather}
S_{1} = LeakyReLU \left( H_{DC1} \left ( F_{i} \right)\right)
\label{eq:SA_1}\\
S_{2} = LeakyReLU \left( H_{DC2} \left ( F_{i} \right ) + S_{1} \right)
\label{eq:SA_2}\\
S = \big[ S_{1} , S_{2} \big]
\label{eq:SA_3}\\
S_{1} = LeakyReLU \left( H_{DC1} \left ( S \right ) \right)
\label{eq:SA_1_1}\\
S_{2} =  LeakyReLU \left( H_{DC2} \left ( S \right ) + S_{1} \right)
\label{eq:SA_2_1}\\
 H_{SA} \left  (  F_{i+1} \right ) = H_{1\times1} * \big[ S_{1} , S_{2} \big] 
 \label{eq:SA_3_1}
\end{gather}

\noindent where $H_{DC1}$ and $H_{DC2}$ denotes the convolution layers with dilation factors 1 and 2 respectively. First, we concatenate the output of two convolution layers to increase the channel size and at the end, we use $1\times1$ convolution to reduce the channels. Thus, our input and output have the same number of channels. Our SA architecture inspires from \cite{haris2018deep} which has shown that upsampling and downsampling module within the architecture improves the accuracy in SR. For the activation unit, by following ref~\cite{lai2017laplacian, tan2018feature}, we prefer the LeakyReLU over ReLU activation whereas we use the linear bottleneck layer as suggested in \cite{sandler2018mobilenetv2}.

\subsubsection{Channel Attention (CA)}

The channel attention (CA) mechanism achieves a lot of success in image classification \cite{hu2018squeeze}. In SISR, RCAN \cite{zhang2018image} introduces the CA layer in the network. CA plays an important role in exploiting the interchannel dependencies because some of them have trivial information while others have the most valuable information. Therefore, we decide to use channel-wise features and incorporate the CA mechanism with SA module in our HRAB. Thus, by following \cite{hu2018squeeze, zhang2018image}, we use the global pooling average to consider the channel-wise global information. We also experiment with global pooling variance as we thought global variance could extract more high-frequencies, in contrast, we get poor results as compare with global pooling average.

Suppose if we have C channels in the feature map $\left [ x_1,x_2, \cdots ,x_C \right ]$ then we can express each `c' feature map as a single value.
\begin{align}
\begin{split}
\label{eq:GP}
z_{c}\left ( x_{c} \right )=\frac{1}{H\times W}\sum_{i=1}^{H}\sum_{j=1}^{W}x_{c}\left ( i,j \right ),
\end{split}
\end{align} 
here $x_{c}$ is the spatial position $\left ( i,j \right )$ of the feature maps.

To extract the channel-wise dependencies, we use the similar sigmoid gating mechanism as \cite{hu2018squeeze, zhang2018image}. Alike SA, here, we replace the ReLU with LeakyReLU activation.

\vspace{-3mm}
\begin{align}
\begin{split}
\label{eq:CA}
H_{CA} \left  ( F_{i+1}\right ) = f \left ( W_{U} LR \left ( W_{D}z \right ) \right ),
\end{split}
\end{align}

Here $LR\left ( \cdot \right )$ and $f\left ( \cdot \right )$ represent the LeakyReLU and sigmoid gating function respectively whereas $W_{D}$ and $W_{U}$ respectively denote the weights of downscaling and upscaling convolutions. It is noted that it is channel-wise downscaling and upscaling with reduction ratio r.

\begin{table}[ht]
\centering
\caption{Investigation of HRAB module (with and without CA). We examine the best PSNR (dB) on Urban100 (2×) with same training settings.}
\label{tab:HRAB}

\begin{adjustbox}{width=\columnwidth}

\begin{tabular}{@{}|c|c|@{}}
\toprule
Modules                                                                                                                 & SSIM / PSNR                     \\ \midrule
\multirow{2}{*}{\begin{tabular}[c]{@{}c@{}}Spatial attention without \\ channel attention (SA)\end{tabular}}            & \multirow{2}{*}{32.77 / 0.9343} \\
                                                                                                                        &                                 \\ \midrule
\multirow{2}{*}{\begin{tabular}[c]{@{}c@{}}Spatial attention with \\ channel attention (Hybrid Attention)\end{tabular}} & \multirow{2}{*}{32.95 / 0.9357} \\
                                                                                                                        &                                 \\ \bottomrule
\end{tabular}
\end{adjustbox}
\end{table}


\begin{table}[ht]
\centering
\caption{BFF vs HFF structures. We examine the best PSNR (dB) on Urban100 $ \left ( 2 \times \right )$ with same training settings.}
\label{tab:BFFS}
\begin{tabular}{@{}|c|c|@{}}
\toprule
Method                          & PSNR / SSIM    \\ \midrule
MSRN with HFF {[}15{]}          & 32.22 / 0.9326 \\ \midrule
MSRN {[}15{]} with proposed BFF & 32.44 / 0.9315 \\ \midrule
Our HRAN with HFF               & 32.69 / 0.9334 \\ \midrule
Our HRAN with proposed BFF      & 32.95 / 0.9375 \\ \bottomrule
\end{tabular}
\end{table}

\subsection{Implementation details}

For training the HRAN network, we employ 4 RG blocks in our main architecture and in each RG block, there are 8 HRAB modules which are followed by $3\times3$ convolution. For the dilated convolution layers, we use the $3\times3$ convolution with dilation factor 1 and 2. We use $C=64$ filters in all the layers except the final layer which has 3 filters to produce a color image though our network can work for both gray and color images. For the channel-downscaling in CA mechanism, we set a reduction factor $r=4$.


\begin{table*}[t]
\caption{Quantitative Comparisons of state-of-the-art methods for BI degradation model. Best, 2nd best and 3rd best results are respectively shown with \textcolor{magenta}{Magenta}, \textcolor{blue}{Blue}, and \textcolor{green}{Green} colors. }
\centering
\label{tab:results_psnr_ssim_x2348}

\begin{center}
\scalebox{0.8}{
\begin{tabular}{|l|c|c|c|c|c|c|c|c|c|c|c|}
\hline
\multirow{2}{*}{Method} & \multirow{2}{*}{Scale} &  \multicolumn{2}{c|}{Set5} &  \multicolumn{2}{c|}{Set14} &  \multicolumn{2}{c|}{B100} &  \multicolumn{2}{c|}{Urban100} &  \multicolumn{2}{c|}{Manga109}  
\\
\cline{3-12}
&  & PSNR & SSIM & PSNR & SSIM & PSNR & SSIM & PSNR & SSIM & PSNR & SSIM 
\\
\hline
\hline
Bicubic & $\times$2 
& 33.66
 & 0.9299
  & 30.24
   & 0.8688
    & 29.56
     & 0.8431
      & 26.88
       & 0.8403
        & 30.80
         & 0.9339
                  
\\
SRCNN~\cite{dong2016image} & $\times$2 
& 36.66
 & 0.9542
  & 32.45
   & 0.9067
    & 31.36
     & 0.8879
      & 29.50
       & 0.8946
        & 35.60
         & 0.9663
                   
\\
FSRCNN~\cite{dong2016accelerating} & $\times$2 
& 37.05
 & 0.9560
  & 32.66
   & 0.9090
    & 31.53
     & 0.8920
      & 29.88
       & 0.9020
        & 36.67
         & 0.9710
                   
\\
VDSR~\cite{kim2016accurate} & $\times$2 
& 37.53
 & 0.9590
  & 33.05
   & 0.9130
    & 31.90
     & 0.8960
      & 30.77
       & 0.9140
        & 37.22
         & 0.9750
                   
\\
LapSRN~\cite{lai2017laplacian} & $\times$2 
& 37.52
 & 0.9591
  & 33.08
   & 0.9130
    & 31.08
     & 0.8950
      & 30.41
       & 0.9101
        & 37.27
         & 0.9740
                   
\\
MemNet~\cite{tai2017memnet} & $\times$2 
& 37.78
 & 0.9597
  & 33.28
   & 0.9142
    & 32.08
     & 0.8978
      & 31.31
       & 0.9195
        & 37.72
         & 0.9740
                   
\\
EDSR~\cite{lim2017enhanced} & $\times$2 
& 38.11
 & 0.9602
  & \textcolor{green}{33.92}
   & 0.9195
    & \textcolor{green}{32.32}
     & \textcolor{green}{0.9013}
      & \textcolor{green}{32.93}
       & 0.9351
        & -/-
         & -/-
                   
\\
SRMDNF~\cite{zhang2018learning} & $\times$2 
& 37.79
 & 0.9601
  & 33.32
   & 0.9159
    & 32.05
     & 0.8985
      & 31.33
       & 0.9204
        & 38.07
         & 0.9761
    
\\
RDN~\cite{zhang2018residual} & $\times$2 
& \textcolor{blue}{38.24}
 & \textcolor{magenta}{0.9614}
  & \textcolor{magenta}{34.01}
   & \textcolor{magenta}{0.9212}
    & \textcolor{blue}{32.34}
     & \textcolor{blue}{0.9017}
      & 32.89
       & \textcolor{green}{0.9353}
        & \textcolor{blue}{39.18}
         & \textcolor{green}{0.9780}
         
\\

DCSR~\cite{zhang2019dcsr} & $\times$2 
& 37.54
 & 0.9587
  & 33.14
   & 0.9141
    & 31.90
     & 0.8959
      & 30.76
       & 0.9142
        & -/-
         & -/-
         
\\

MSRN~\cite{Li_2018_ECCV} & $\times$2 
& 38.08
 & 0.9605
  & 33.74
   & 0.9170
    & 32.23
     & 0.9013
      & 32.22
       & 0.9326
        & 38.82
         & \textcolor{magenta}{0.9868}
         
\\
HRAN (ours) & $\times$2 
& \textcolor{green}{38.21}
 & \textcolor{blue}{0.9613}
  & 33.85
   & \textcolor{green}{0.9200}
    & \textcolor{blue}{32.34}
     & \textcolor{green}{0.9016}
      & \textcolor{blue}{32.95}
       & \textcolor{blue}{0.9357}
        & \textcolor{green}{39.12}
         & \textcolor{green}{0.9780}

\\
HRAN+ (ours) & $\times$2 
& \textcolor{magenta}{38.25}
 & \textcolor{magenta}{0.9614}
  & \textcolor{blue}{33.99}
   & \textcolor{blue}{0.9211}
    & \textcolor{magenta}{32.38}
     & \textcolor{magenta}{0.9020}
      & \textcolor{magenta}{33.12}
       & \textcolor{magenta}{0.9370}
        & \textcolor{magenta}{39.29}
         & \textcolor{blue}{0.9785}

\\
\hline
\hline
Bicubic & $\times$3 
& 30.39
 & 0.8682
  & 27.55
   & 0.7742
    & 27.21
     & 0.7385
      & 24.46
       & 0.7349
        & 26.95
         & 0.8556
                  
\\
SRCNN~\cite{dong2016image} & $\times$3
& 32.75
 & 0.9090
  & 29.30
   & 0.8215
    & 28.41
     & 0.7863
      & 26.24
       & 0.7989
        & 30.48
         & 0.9117
                    
\\
FSRCNN~\cite{dong2016accelerating} & $\times$3 
& 33.18
 & 0.9140
  & 29.37
   & 0.8240
    & 28.53
     & 0.7910
      & 26.43
       & 0.8080
        & 31.10
         & 0.9210
                   
\\
VDSR~\cite{kim2016accurate} & $\times$3 
& 33.67
 & 0.9210
  & 29.78
   & 0.8320
    & 28.83
     & 0.7990
      & 27.14
       & 0.8290
        & 32.01
         & 0.9340
                   
\\
LapSRN~\cite{lai2017laplacian} & $\times$3 
& 33.82
 & 0.9227
  & 29.87
   & 0.8320
    & 28.82
     & 0.7980
      & 27.07
       & 0.8280
        & 32.21
         & 0.9350
                   
\\
MemNet~\cite{tai2017memnet} & $\times$3 
& 34.09
 & 0.9248
  & 30.00
   & 0.8350
    & 28.96
     & 0.8001
      & 27.56
       & 0.8376
        & 32.51
         & 0.9369
                   
\\
EDSR~\cite{lim2017enhanced} & $\times$3 
& 34.65
 & 0.9280
  & 30.52
   & 0.8462
    & \textcolor{green}{29.25}
     & \textcolor{blue}{0.8093}
      & \textcolor{blue}{28.80}
       & \textcolor{blue}{0.8653}
        & -/-
         & -/-
                   
\\
SRMDNF~\cite{zhang2018learning} & $\times$3 
& 34.12
 & 0.9254
  & 30.04
   & 0.8382
    & 28.97
     & 0.8025
      & 27.57
       & 0.8398
        & 33.00
         & 0.9403
                   
\\
RDN~\cite{zhang2018residual} & $\times$3 
& \textcolor{blue}{34.71}
 & \textcolor{blue}{0.9296}
  & \textcolor{blue}{30.57}
   & \textcolor{blue}{0.8468}
    & \textcolor{blue}{29.26}
     & \textcolor{blue}{0.8093}
      & \textcolor{blue}{28.80}
       & \textcolor{blue}{0.8653}
        & \textcolor{blue}{34.13}
         & \textcolor{blue}{0.9484}
\\
DCSR~\cite{zhang2019dcsr} & $\times$3 
& 33.94
 & 0.9234
  & 30.28
   & 0.8354
    & 28.86
     & 0.7985
      & 27.24
       & 0.8308
        & -/-
         & -/-
         
\\
MSRN~\cite{Li_2018_ECCV} & $\times$3 
& 34.38
 & 0.9262
  & 30.34
   & 0.8395
    & 29.08
     & 0.8041
      & 28.08
       & 0.8554
        & 33.44
         & 0.9427
         
\\
HRAN (ours) & $\times$3 
& \textcolor{green}{34.69}
 & \textcolor{green}{0.9292}
  & \textcolor{green}{30.54}
   & \textcolor{green}{0.8463}
    & \textcolor{green}{29.25}
     & \textcolor{green}{0.8089}
      & \textcolor{green}{28.76}
       & \textcolor{green}{0.8645}
        & \textcolor{green}{34.08}
         & \textcolor{green}{0.9479}

\\
HRAN+ (ours) & $\times$3 
& \textcolor{magenta}{34.75}
 & \textcolor{magenta}{0.9298}
  & \textcolor{magenta}{30.60}
   & \textcolor{magenta}{0.8474}
    & \textcolor{magenta}{29.29}
     & \textcolor{magenta}{0.8098}
      & \textcolor{magenta}{28.96}
       & \textcolor{magenta}{0.8670}
        & \textcolor{magenta}{34.36}
         & \textcolor{magenta}{0.9492}
         
\\
\hline
\hline
Bicubic & $\times$4 
& 28.42
 & 0.8104
  & 26.00
   & 0.7027
    & 25.96
     & 0.6675
      & 23.14
       & 0.6577
        & 24.89
         & 0.7866
                  
\\
SRCNN~\cite{dong2016image} & $\times$4 
& 30.48
 & 0.8628
  & 27.50
   & 0.7513
    & 26.90
     & 0.7101
      & 24.52
       & 0.7221
        & 27.58
         & 0.8555
                   
\\
FSRCNN~\cite{dong2016accelerating} & $\times$4 
& 30.72
 & 0.8660
  & 27.61
   & 0.7550
    & 26.98
     & 0.7150
      & 24.62
       & 0.7280
        & 27.90
         & 0.8610
                   
\\
VDSR~\cite{kim2016accurate} & $\times$4 
& 31.35
 & 0.8830
  & 28.02
   & 0.7680
    & 27.29
     & 0.0726
      & 25.18
       & 0.7540
        & 28.83
         & 0.8870
                   
\\
LapSRN~\cite{lai2017laplacian} & $\times$4 
& 31.54
 & 0.8850
  & 28.19
   & 0.7720
    & 27.32
     & 0.7270
      & 25.21
       & 0.7560
        & 29.09
         & 0.8900
                   
\\
MemNet~\cite{tai2017memnet} & $\times$4 
& 31.74
 & 0.8893
  & 28.26
   & 0.7723
    & 27.40
     & 0.7281
      & 25.50
       & 0.7630
        & 29.42
         & 0.8942
                   
\\
EDSR~\cite{lim2017enhanced} & $\times$4 
& \textcolor{green}{32.46}
 & 0.8968
  & 28.80
   & 0.7876
    & 27.71
     & 0.7420
      & 26.64
       & 0.8033
        & -/-
         & -/-
                   
\\
SRMDNF~\cite{zhang2018learning} & $\times$4 
& 31.96
 & 0.8925
  & 28.35
   & 0.7787
    & 27.49
     & 0.7337
      & 25.68
       & 0.7731
        & 30.09
         & 0.9024

\\
RDN~\cite{zhang2018residual} & $\times$4 
& \textcolor{blue}{32.47}
 & \textcolor{blue}{0.8990}
  & \textcolor{blue}{28.81}
   & \textcolor{blue}{0.7871}
    & \textcolor{blue}{27.72}
     & \textcolor{blue}{0.7419}
      & \textcolor{green}{26.61}
       & \textcolor{green}{0.8028}
        & \textcolor{blue}{31.00}
         & \textcolor{blue}{0.9151}

\\         
DCSR~\cite{zhang2019dcsr} & $\times$4 
& 31.58
 & 0.8870
  & 28.21
   & 0.7715
    & 27.32
     & 0.7264
      & \textcolor{magenta}{27.24}
       & \textcolor{magenta}{0.8308}
        & -/- 
         & -/-
         
\\
MSRN~\cite{Li_2018_ECCV} & $\times$4 
& 32.07
 & 0.8903
  & 28.60
   & 0.775
    & 27.52
     & 0.7273
      & 26.04
       & 0.7896
        & 30.17
         & 0.9034
         
\\
HRAN (ours) & $\times$4 
& 32.43
 & \textcolor{green}{0.8976}
  & \textcolor{green}{28.76}
   & \textcolor{green}{0.7863}
    & \textcolor{green}{27.70}
     & 0.7407
      & 26.55
       & 0.8006
        & \textcolor{green}{30.94}
         &\textcolor{green}{0.9143}

\\
HRAN+ (ours) & $\times$4 
& \textcolor{magenta}{32.56}
 & \textcolor{magenta}{0.8991}
  & \textcolor{magenta}{28.86}
   & \textcolor{magenta}{0.7880}
    & \textcolor{magenta}{27.76}
     & \textcolor{magenta}{0.7420}
      & \textcolor{blue}{26.74}
       & \textcolor{blue}{0.8046}
        & \textcolor{magenta}{31.26}
         & \textcolor{magenta}{0.9172}
              
\\
\hline
\hline
Bicubic & $\times$8 
& 24.40
 & 0.6580
  & 23.10
   & 0.5660
    & 23.67
     & 0.5480
      & 20.74
       & 0.5160
        & 21.47
         & 0.6500
                 
\\
SRCNN~\cite{dong2016image} & $\times$8 
& 25.33
 & 0.6900
  & 23.76
   & 0.5910
    & 24.13
     & 0.5660
      & 21.29
       & 0.5440
        & 22.46
         & 0.6950
                   
\\
FSRCNN~\cite{dong2016accelerating} & $\times$8 
& 20.13
 & 0.5520
  & 19.75
   & 0.4820
    & 24.21
     & 0.5680
      & 21.32
       & 0.5380
        & 22.39
         & 0.6730
                   
\\
SCN~\cite{wang2015deep} & $\times$8 
& 25.59
 & 0.7071
  & 24.02
   & 0.6028
    & 24.30
     & 0.5698
      & 21.52
       & 0.5571
        & 22.68
         & 0.6963

\\
VDSR~\cite{kim2016accurate} & $\times$8 
& 25.93
 & 0.7240
  & 24.26
   & 0.6140
    & 24.49
     & 0.5830
      & 21.70
       & 0.5710
        & 23.16
         & 0.7250
                   
\\   
LapSRN~\cite{lai2017laplacian} & $\times$8 
& 26.15
 & 0.7380
  & 24.35
   & 0.6200
    & 24.54
     & 0.5860
      & 21.81
       & 0.5810
        & 23.39
         & 0.7350
                   
\\
MemNet~\cite{tai2017memnet} & $\times$8 
& 26.16
 & 0.7414
  & 24.38
   & 0.6199
    & 24.58
     & 0.5842
      & 21.89
       & 0.5825
        & 23.56
         & 0.7387

\\
EDSR~\cite{lim2017enhanced} & $\times$8 
& \textcolor{green}{26.96}
 & \textcolor{green}{0.7762}
  & \textcolor{green}{24.91}
   & \textcolor{blue}{0.6420}
    & \textcolor{green}{24.81}
     & \textcolor{blue}{0.5985}
      & \textcolor{green}{22.51}
       & \textcolor{green}{0.6221}
        & -/-
         & -/-

\\
MSRN~\cite{Li_2018_ECCV} & $\times$8 
& 26.59
 & 0.7254
  & 24.88
   & 0.5961
    & 24.70
     & 0.5410
      & 22.37
       & 0.5977
        & \textcolor{green}{24.28}
         & \textcolor{green}{0.7517}         
\\
HRAN (ours) & $\times$8 
& \textcolor{blue}{27.11}
 & \textcolor{blue}{0.7798}
  & \textcolor{blue}{25.01}
   & \textcolor{green}{0.6419}
    & \textcolor{blue}{24.83}
     & \textcolor{green}{0.5983}
      & \textcolor{blue}{22.57}
       & \textcolor{blue}{0.6223}
        & \textcolor{blue}{24.64}
         & \textcolor{blue}{0.7817}

\\
HRAN+ (ours) & $\times$8 
& \textcolor{magenta}{27.18}
 & \textcolor{magenta}{0.7828}
  & \textcolor{magenta}{25.12}
   & \textcolor{magenta}{0.6450}
    & \textcolor{magenta}{24.89}
     & \textcolor{magenta}{0.6001}
      & \textcolor{magenta}{22.73}
       & \textcolor{magenta}{0.6280}
        & \textcolor{magenta}{24.87}
         & \textcolor{magenta}{0.7878}
           
\\
\hline             
\end{tabular}}
\end{center}

\end{table*}

%

%



\begin{figure*}[tp]
	\newlength\fsdurthree
	\setlength{\fsdurthree}{-1.5mm}
	\scriptsize
	\centering
	\begin{tabular}{cc}
		\begin{adjustbox}{valign=t}
		\tiny
			\begin{tabular}{c}
				\includegraphics[width=0.170\textwidth]{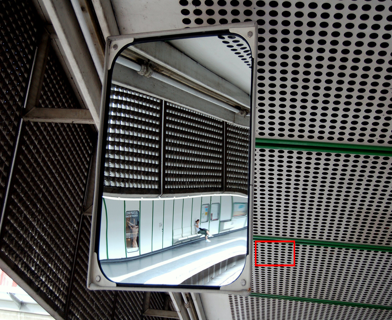}
				\\
				Urban100 ($4\times$):
				\\
				img\_004
			\end{tabular}
		\end{adjustbox}
		\hspace{-2.3mm}
		\begin{adjustbox}{valign=t}
		\tiny
			\begin{tabular}{cccccc}
				\includegraphics[width=\widthscalefive \textwidth]{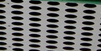} \hspace{\fsdurthree} &
				\includegraphics[width=\widthscalefive \textwidth]{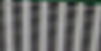} \hspace{\fsdurthree} &
				\includegraphics[width=\widthscalefive \textwidth]{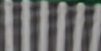} \hspace{\fsdurthree} &
				\includegraphics[width=\widthscalefive \textwidth]{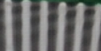} \hspace{\fsdurthree} &
				\includegraphics[width=\widthscalefive \textwidth]{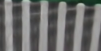} 
				\\
				HR \hspace{\fsdurthree} &
				Bicubic \hspace{\fsdurthree} &
				SRCNN~\cite{dong2016image} \hspace{\fsdurthree} &
				FSRCNN~\cite{dong2016accelerating} \hspace{\fsdurthree} &
				VDSR~\cite{kim2016accurate} 
				
				\\
				\includegraphics[width=\widthscalefive \textwidth]{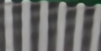} \hspace{\fsdurthree} &
				\includegraphics[width=\widthscalefive \textwidth]{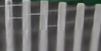} \hspace{\fsdurthree} &
				\includegraphics[width=\widthscalefive \textwidth]{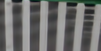} \hspace{\fsdurthree} &
				\includegraphics[width=\widthscalefive \textwidth]{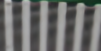} \hspace{\fsdurthree} &
				\includegraphics[width=\widthscalefive \textwidth]{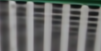} 
				\\ 
				LapSRN~\cite{lai2017laplacian} \hspace{\fsdurthree} &
				MemNet~\cite{tai2017memnet} \hspace{\fsdurthree} &
				EDSR~\cite{lim2017enhanced} \hspace{\fsdurthree} &
				SRMDNF~\cite{zhang2018learning} \hspace{\fsdurthree} &
				HRAN 
				
				\\
			\end{tabular}
		\end{adjustbox}
		\vspace{0.5mm}
		\\
		\begin{adjustbox}{valign=t}
		\tiny
			\begin{tabular}{c}
				\includegraphics[width=0.170\textwidth]{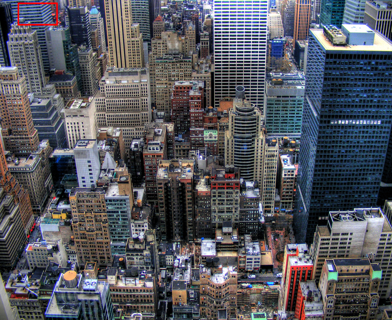}
				\\
				Urban100 ($4\times$):
				\\
				img\_073
			\end{tabular}
		\end{adjustbox}
		\hspace{-2.3mm}
		\begin{adjustbox}{valign=t}
		\tiny
			\begin{tabular}{cccccc}
				\includegraphics[width=\widthscalefive \textwidth]{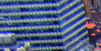} \hspace{\fsdurthree} &
				\includegraphics[width=\widthscalefive \textwidth]{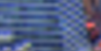} \hspace{\fsdurthree} &
				\includegraphics[width=\widthscalefive \textwidth]{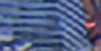} \hspace{\fsdurthree} &
				\includegraphics[width=\widthscalefive \textwidth]{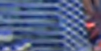} \hspace{\fsdurthree} &
				\includegraphics[width=\widthscalefive \textwidth]{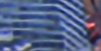} 
				\\
				HR \hspace{\fsdurthree} &
				Bicubic \hspace{\fsdurthree} &
				SRCNN~\cite{dong2016image} \hspace{\fsdurthree} &
				FSRCNN~\cite{dong2016accelerating} \hspace{\fsdurthree} &
				VDSR~\cite{kim2016accurate} 
				
				\\
				\includegraphics[width=\widthscalefive \textwidth]{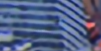} \hspace{\fsdurthree} &
				\includegraphics[width=\widthscalefive \textwidth]{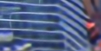} \hspace{\fsdurthree} &
				\includegraphics[width=\widthscalefive \textwidth]{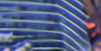} \hspace{\fsdurthree} &
				\includegraphics[width=\widthscalefive \textwidth]{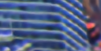} \hspace{\fsdurthree} &
				\includegraphics[width=\widthscalefive \textwidth]{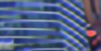}  
				\\ 
				LapSRN~\cite{lai2017laplacian} \hspace{\fsdurthree} &
				MemNet~\cite{tai2017memnet} \hspace{\fsdurthree} &
				EDSR~\cite{lim2017enhanced} \hspace{\fsdurthree} &
				SRMDNF~\cite{zhang2018learning} \hspace{\fsdurthree} &
				HRAN 
				
				\\
			\end{tabular}
		\end{adjustbox}
		\vspace{0.5mm}
		\\		
		\begin{adjustbox}{valign=t}
		\tiny
			\begin{tabular}{c}
				\includegraphics[width=0.170\textwidth]{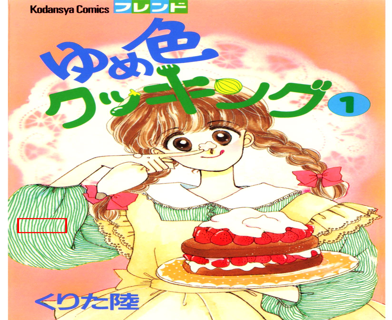}
				\\
				Manga109 ($4\times$):
				\\
				YumeiroCooking
			\end{tabular}
		\end{adjustbox}
		\hspace{-2.3mm}
		\begin{adjustbox}{valign=t}
		\tiny
			\begin{tabular}{cccccc}
				\includegraphics[width=\widthscalefive \textwidth]{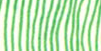} \hspace{\fsdurthree} &
				\includegraphics[width=\widthscalefive \textwidth]{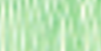} \hspace{\fsdurthree} &
				\includegraphics[width=\widthscalefive \textwidth]{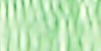} \hspace{\fsdurthree} &
				\includegraphics[width=\widthscalefive \textwidth]{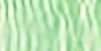} \hspace{\fsdurthree} &
				\includegraphics[width=\widthscalefive \textwidth]{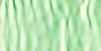} 
				\\
				HR \hspace{\fsdurthree} &
				Bicubic \hspace{\fsdurthree} &
				SRCNN~\cite{dong2016image} \hspace{\fsdurthree} &
				FSRCNN~\cite{dong2016accelerating} \hspace{\fsdurthree} &
				VDSR~\cite{kim2016accurate} 
				
				\\
				\includegraphics[width=\widthscalefive \textwidth]{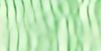} \hspace{\fsdurthree} &
				\includegraphics[width=\widthscalefive \textwidth]{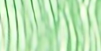} \hspace{\fsdurthree} &
				\includegraphics[width=\widthscalefive \textwidth]{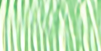} \hspace{\fsdurthree} &
				\includegraphics[width=\widthscalefive \textwidth]{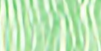} \hspace{\fsdurthree} &
				\includegraphics[width=\widthscalefive \textwidth]{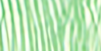}  
				\\ 
				LapSRN~\cite{lai2017laplacian} \hspace{\fsdurthree} &
				MemNet~\cite{tai2017memnet} \hspace{\fsdurthree} &
				EDSR~\cite{lim2017enhanced} \hspace{\fsdurthree} &
				SRMDNF~\cite{zhang2018learning} \hspace{\fsdurthree} &
				HRAN 
				
				\\
			\end{tabular}
		\end{adjustbox}
	\vspace{3mm}
	\end{tabular}
	\caption{
		Qualitative results for $4\times$ SR with BI model on Urban100 and Manga109 datasets.}
	
\label{fig:result_4x_Urban100_Manga109}
\end{figure*}



\section{Experimental Results}
In this section, we explain the experimental analysis of our method. For this purpose, we use several public datasets that are considered as the benchmark in SISR. We provide the results of both the quantitative and qualitative experiments for the comparison of our method with several state-of-the-art networks. For the datasets, we follow the recent trends \cite{lim2017enhanced, zhang2018learning, zhang2018residual, Li_2018_ECCV, timofte2017ntire} and use DIV2K dataset as the training set, since it contains the high-resolution images. For testing, we choose widely used standard datasets: Set5 \cite{bevilacqua2012low}, Set14\cite{zeyde2010single}, BDS100 \cite{martin2001database}, Urban100 \cite{huang2015single} and Manga109 \cite{matsui2017sketch}. For the degradation, we use the Bicubic Interpolation (BI). 

We evaluate our results with peak signal-to-noise ratio (PSNR) and structural similarity (SSIM) \cite{wang2004image} on luminance channel \ie, Y of transformed YCbCr space and we remove P-pixels from each border (P refers to upscaling factor). We provide the results for scaling factor $\times2$, $\times3$, $\times4$, and $\times8$.


For the training settings, we follow the settings in \cite{Li_2018_ECCV}. We extract 16 $LR$ patches randomly in each training batch with the size of 64$\times$64. We use ADAM optimizer with learning rate $lr=10^{-4}$ which decreases to half after every $2\times10^{5}$ iterations of back-propagation. We use PyTorch framework to implement our models with GeForce RTX 2080 Ti GPU. 


\subsection{Ablation studies}
We conduct a series of ablation studies to show the effectiveness of our model. In the first experiment, we train our model with and without CA and compare their performance with our HRAB module. For the training, we use Urban100 dataset\cite{huang2015single} as it consists of large dataset. The results are shown in Table~\ref{tab:HRAB}. We observe that our SA module alone achieves 32.77 dB PSNR. We also experimented on CA module alone though results were unsatisfactory. Whereas, when we combine SA with CA, \ie, our HRAB module, it achieves the 32.95 db PSNR. This study suggests we need HRAB module containing both the spatial and channel attention for accurate SR results. We also investigate about our BFF structure using HRAB module and tested the both BFF and HFF on MSRN \cite{Li_2018_ECCV} and our proposed HRAN model to verify the effectiveness of BFF on both models. It is evident from the results that BFF structure improves the PSNR of MSRN \cite{Li_2018_ECCV} from 32.22 dB to 32.44 dB with BFF. Moreover, proposed HRAN and BFF together significantly increase the accuracy which show the effectiveness of our BFF structure.




\begin{figure}[bp]
\scriptsize
\centering
\includegraphics[width = 80mm, height = 40mm]{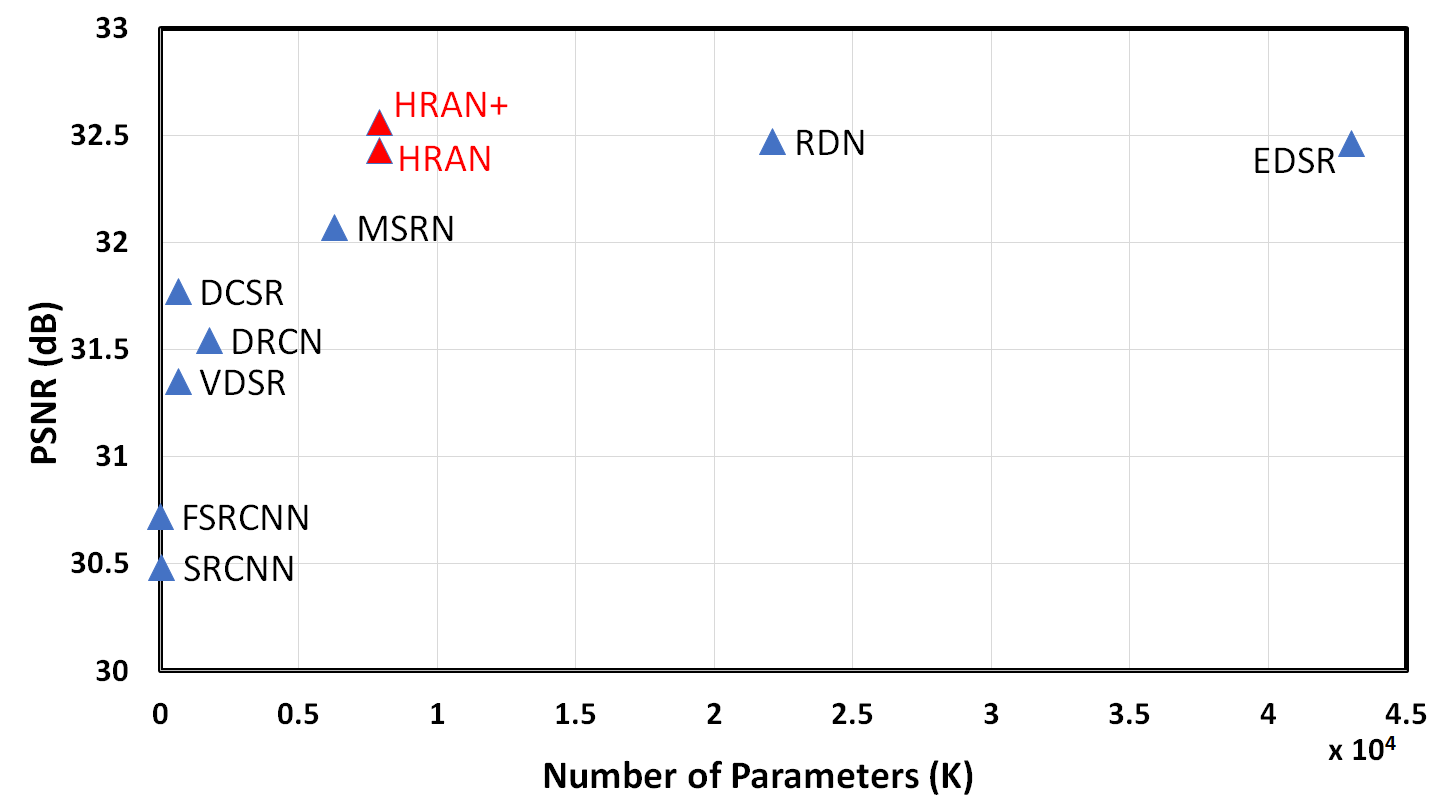}

\caption{Comparison of memory and performance. Results are evaluated on Set5 ($\times4$).}  
\label{fig:memory_comparison}
\end{figure}

\subsection{Comparison with State-of-the-art Methods}
We compare our method with 10 state-of-the-art SISR methods: SRCNN \cite{dong2016image}, FSRCNN\cite{dong2016accelerating}, VDSR \cite{kim2016accurate}, LapSRN \cite{lai2017laplacian}, MEMNet \cite{tai2017memnet}, EDSR \cite{lim2017enhanced} , SRMDNF \cite{zhang2018learning}, RDN \cite{zhang2018residual}, DCSR \cite{zhang2019dcsr} and MSRN~\cite{Li_2018_ECCV}. By following \cite{lim2017enhanced, Timofte_2016_CVPR}, , we also use self-ensemble strategy to improve the accuracy of our model at test time. 

We show our quantitative evaluation results in Table~\ref{tab:results_psnr_ssim_x2348} for the scale factor of $\times2$, $\times3$, $\times4$, and $\times8$. It is evident from the results that our method outperforms most of the previous methods. Our self-ensemble model achieves the highest PSNR amongst all the models. Although RDN \cite{zhang2018residual} has shown slightly better performance, from Figure~\ref{fig:memory_comparison}, we observe that RDN \cite{zhang2018residual} has about 22M parameters, in contrast, our HRAN model has only 7.94 M parameters though our model shows comparable performance. Instead of increasing the depth and dense connections, our HRAN model with HRAB and BFF detect the deep features without increasing the depth of the network. Hence, this observation indicates that we can improve the network performance with HRAB and RG along with BFF without increasing the network depth. This also suggests that our network can further improve the accuracy with more HRAB's and RG's, though, we aim to achieve the greater accuracy by considering the memory computations.   

Moreover, we present the qualitative results in Figure~\ref{fig:result_4x_Urban100_Manga109}. The results of other methods are derived from \cite{zhang2018image}. In Figure~\ref{fig:result_4x_Urban100_Manga109}, it can be observed from `img\_004' image our HRAN method recovers the lattices in more details, meanwhile, other methods experience the blurring artifacts. Similar behavior is also observed in `Yumeiro-Cooking' image where other methods produce blurry lines and our HRAN produces the lines similar to HR image. It shows that our model reconstructs the fine details in output SR image through extracted deep features with RGs which are then efficiently utilized by BFF. 


\subsection{Model Complexity Analysis}
Since, we are targeting the maximum accuracy with limited memory computation, therefore our performance is best visible when we see the Table.\ref{tab:results_psnr_ssim_x2348} along with Figure.~\ref{fig:memory_comparison}. In Figure.~\ref{fig:memory_comparison}, we compare our model size and its performance on Set5 \cite{bevilacqua2012low} ($\times$4). As we can observe that our HRAN model has fewer parameters compared to RDN \cite{zhang2018residual} and EDSR \cite{lim2017enhanced}, it still achieves the comparable performance whereas our HRAN+ outperforms the state-of-the-art methods. We have also shown analysis on much larger scale ($\times$8) in supplementary materials. These results demonstrate the effective utilization of the features that result in performance gain in SISR.



\section{Conclusions}
In this paper, we propose a hybrid residual attention network (HRAN) to detect the most informative multiscale spatial features for the accurate SR image. Proposed hybrid residual attention block (HRAB) module fully utilize the high-frequency information from input features with a combination of the spatial attention (SA) and channel attention (CA). In addition, the binarized feature fusion (BFF) structure allows us to smoothly transmit all the features at the end of the network for reconstruction. Furthermore, we propose to adopt the global, short and long skip connection and residual groups (RG) to ease the flow of information. Our comprehensive experiments show the efficacy of the proposed model.

\newpage
\clearpage 

\section*{\centering Supplementary Material}
\setcounter{section}{0}


   In this supplementary submission, we present more qualitative results with the different scaling factors. Furthermore, we also compare our method's computation complexity with the large scaling factor.


\section{Experimental Results}

\subsection{Model Complexity Analysis}

\begin{figure}[pb]
\scriptsize
\centering
\includegraphics[width = 80mm, height = 40mm]{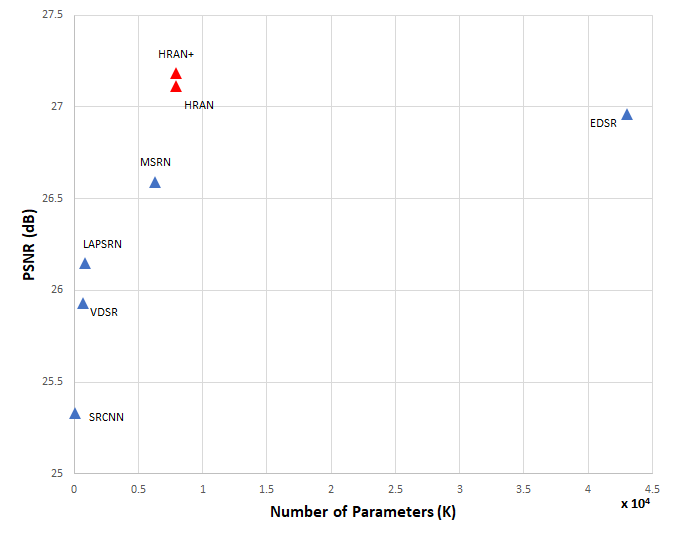}

\caption{Comparison of memory and performance. Results are evaluated on Set5 ($\times8$).}  
\label{fig:memory_comparison_x8}
\end{figure}

When it comes to the large scaling factor, the reconstruction of the SR image becomes more difficult and the SR problem is intensified due to very limited information in the LR image. In this section, we compare our model computation complexity and performance on the large scaling factor ($8\times$) in terms of a number of parameters and peak signal-to-noise ratio (PSNR) respectively in Figure. \ref{fig:memory_comparison_x8}. The results in Figure. \ref{fig:memory_comparison_x8} shows that our HRAN and HRAN+ models outperform all the models including EDSR, and MSRN, for the scaling factor 8$\times$ with the low number of parameters. 

\subsection{Visual Comparisons}

 In this section, we compare our qualitative results with the state-of-the-art methods: SRCNN \cite{dong2016image}, SPMSR \cite{peleg2014statistical}, FSRCNN \cite{dong2016accelerating}, VDSR \cite{kim2016accurate}, IRCNN \cite{zhang2017learning}, SRMDNF \cite{zhang2018learning}, SCN \cite{wang2015deep}, DRRN \cite{Tai-DRRN-2017}, LapSRN \cite{lai2017laplacian}, MSLapSRN \cite{MSLapSRN}, Enet-PAT \cite{sajjadi2017enhancenet}, MemNet \cite{tai2017memnet}, and EDSR \cite{lim2017enhanced}.

\begin{figure*}
	\newlength\fsfourteen
	\setlength{\fsfourteen}{-1.3mm}
	\scriptsize
	\centering
	\begin{tabular}{cc}
	\tiny
		\begin{adjustbox}{valign=t}
			\begin{tabular}{ccccccc}
				\includegraphics[width=0.1\textwidth]{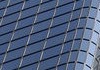} \hspace{\fsfourteen} &
				\includegraphics[width=0.1\textwidth]{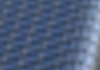}  \hspace{\fsfourteen} &
				\includegraphics[width=0.1\textwidth]{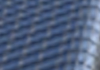} \hspace{\fsfourteen} &
				\includegraphics[width=0.1\textwidth]{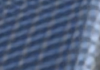} \hspace{\fsfourteen} &
				\includegraphics[width=0.1\textwidth]{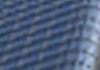} \hspace{\fsfourteen} &
				\includegraphics[width=0.1\textwidth]{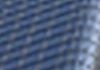}
				\hspace{\fsfourteen} &
				\includegraphics[width=0.1\textwidth]{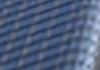}
				\\
				HR \hspace{\fsfourteen} &
				Bicubic \hspace{\fsfourteen} &
				SRCNN~\cite{dong2016image} \hspace{\fsfourteen} &
				FSRCNN~\cite{dong2016accelerating} \hspace{\fsfourteen} &
				SCN~\cite{wang2015deep} \hspace{\fsfourteen} &
				VDSR~\cite{kim2016accurate} &
				DRRN~\cite{Tai-DRRN-2017} \hspace{\fsfourteen} 
				\\
				\includegraphics[width=0.1\textwidth]{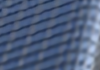} \hspace{\fsfourteen} &
				\includegraphics[width=0.1\textwidth]{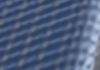} \hspace{\fsfourteen} &
				\includegraphics[width=0.1\textwidth]{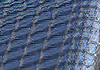} \hspace{\fsfourteen} &
				\includegraphics[width=0.1\textwidth]{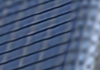} \hspace{\fsfourteen} &
				\includegraphics[width=0.1\textwidth]{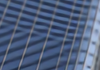} \hspace{\fsfourteen} &
				\includegraphics[width=0.1\textwidth]{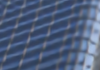} \hspace{\fsfourteen} &
				\includegraphics[width=0.1\textwidth]{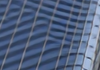} 
				\\ 
				LapSRN~\cite{lai2017laplacian} \hspace{\fsfourteen} &
				MSLapSRN~\cite{MSLapSRN} \hspace{\fsfourteen} &
				ENet-PAT~\cite{sajjadi2017enhancenet} \hspace{\fsfourteen} &
				MemNet~\cite{tai2017memnet} \hspace{\fsfourteen} &
				EDSR~\cite{lim2017enhanced} \hspace{\fsfourteen} &
				SRMDNF~\cite{zhang2018learning} \hspace{\fsfourteen} &
				HRAN (ours)
				\\
			\end{tabular}
		\end{adjustbox}

	\end{tabular}
\vspace{3mm}	
	\caption{
		``img\_074" from Urban100 (4$\times$): State-of-the-art results with Bicubic (BI) degradation.
	}
	\label{fig:result_4x_14}

\end{figure*}

We show the experiments' results with the different scaling factors:  $3\times$, $4\times$, and $8\times$.  In Figure.~\ref{fig:result_4x_14}, we can visualize that most of the methods fail to reconstruct the fine details in `img\_062' and `img\_078' and have blurry effects. Although SRMDNF has recovered the horizontal and vertical lines but output result is more blurry. Whereas, our results have no blurry effect and have shown similar visual performance than EDSR.

\begin{figure*}[thp]
	\scriptsize
	\centering
	\begin{tabular}{cc}
		\begin{adjustbox}{valign=t}
		\tiny
			\begin{tabular}{c}
				\includegraphics[width=0.170\textwidth]{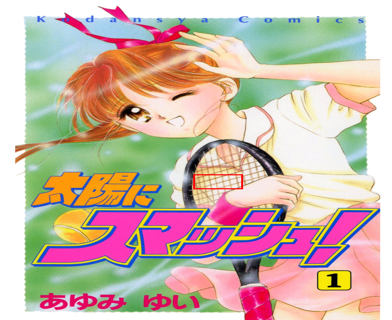}
				\\
				Manga109 ($8\times$)
				\\
				TaiyouNiSmash
			\end{tabular}
		\end{adjustbox}
		\hspace{-2.3mm}
		\begin{adjustbox}{valign=t}
		\tiny
			\begin{tabular}{cccccc}
				\includegraphics[width=\widthscalefive \textwidth]{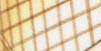} \hspace{\fsdurthree} &
				\includegraphics[width=\widthscalefive \textwidth]{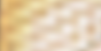} \hspace{\fsdurthree} &
				\includegraphics[width=\widthscalefive \textwidth]{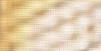} \hspace{\fsdurthree} &
				\includegraphics[width=\widthscalefive \textwidth]{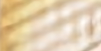} \hspace{\fsdurthree} &
				\includegraphics[width=\widthscalefive \textwidth]{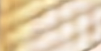} 
				\\
				HR \hspace{\fsdurthree} &
				Bicubic \hspace{\fsdurthree} &
				SRCNN~\cite{dong2016image} \hspace{\fsdurthree} &
				SCN~\cite{wang2015deep} \hspace{\fsdurthree} &
				VDSR~\cite{kim2016accurate} 
				
				\\
				\includegraphics[width=\widthscalefive \textwidth]{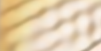} \hspace{\fsdurthree} &
				\includegraphics[width=\widthscalefive \textwidth]{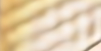} \hspace{\fsdurthree} &
				\includegraphics[width=\widthscalefive \textwidth]{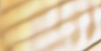} \hspace{\fsdurthree} &
				\includegraphics[width=\widthscalefive \textwidth]{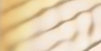} \hspace{\fsdurthree} &
				\includegraphics[width=\widthscalefive \textwidth]{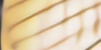} 
				\\ 
				LapSRN~\cite{lai2017laplacian} \hspace{\fsdurthree} &
				MemNet~\cite{tai2017memnet} \hspace{\fsdurthree} &
				EDSR~\cite{lim2017enhanced} \hspace{\fsdurthree} &
				MSLapSRN~\cite{MSLapSRN} \hspace{\fsdurthree} &
				HRAN 
				
				\\
			\end{tabular}
		\end{adjustbox}
		\vspace{0.5mm}
		\\
		\begin{adjustbox}{valign=t}
		\tiny
			\begin{tabular}{c}
				\includegraphics[width=0.170\textwidth]{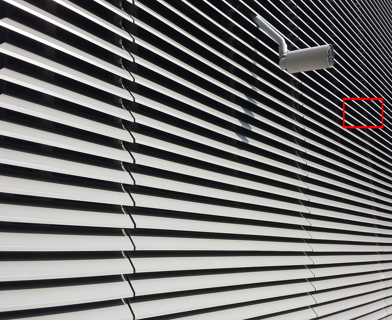}
				\\
				Urban100 ($8\times$):
				\\
				img\_040
			\end{tabular}
		\end{adjustbox}
		\hspace{-2.3mm}
		\begin{adjustbox}{valign=t}
		\tiny
			\begin{tabular}{cccccc}
				\includegraphics[width=\widthscalefive \textwidth]{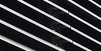} \hspace{\fsdurthree} &
				\includegraphics[width=\widthscalefive \textwidth]{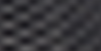} \hspace{\fsdurthree} &
				\includegraphics[width=\widthscalefive \textwidth]{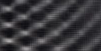} \hspace{\fsdurthree} &
				\includegraphics[width=\widthscalefive \textwidth]{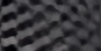} \hspace{\fsdurthree} &
				\includegraphics[width=\widthscalefive \textwidth]{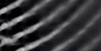} 
				\\
				HR \hspace{\fsdurthree} &
				Bicubic \hspace{\fsdurthree} &
				SRCNN~\cite{dong2016image} \hspace{\fsdurthree} &
				SCN~\cite{wang2015deep} \hspace{\fsdurthree} &
				VDSR~\cite{kim2016accurate} 
				
				\\
				\includegraphics[width=\widthscalefive \textwidth]{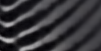} \hspace{\fsdurthree} &
				\includegraphics[width=\widthscalefive \textwidth]{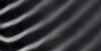} \hspace{\fsdurthree} &
				\includegraphics[width=\widthscalefive \textwidth]{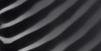} \hspace{\fsdurthree} &
				\includegraphics[width=\widthscalefive \textwidth]{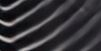} \hspace{\fsdurthree} &
				\includegraphics[width=\widthscalefive \textwidth]{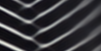}  
				\\ 
				LapSRN~\cite{lai2017laplacian} \hspace{\fsdurthree} &
				MemNet~\cite{tai2017memnet} \hspace{\fsdurthree} &
				MSLapSRN~\cite{MSLapSRN} \hspace{\fsdurthree} &
				EDSR~\cite{lim2017enhanced} \hspace{\fsdurthree} &
				HRAN 
				
				\\
			\end{tabular}
		\end{adjustbox}
		\vspace{0.5mm}

	\vspace{3mm}
	\end{tabular}
	\caption{
		Qualitative results for 8$\times$ SR with BI model on Manga109 and Urban100.}
	
\label{fig:result_8x_Urban100_Manga109}
\end{figure*}

For further illustrations, we also analyze our results on 8$\times$ super-resolution (SR) in Figure~\ref{fig:result_8x_Urban100_Manga109}. When the scaling factor increases, we get very limited details in the LR image.  From the `img\_040' image, what we observe that Bicubic interpolation does not recover the original patterns. Those methods (SRCNN, MemNet, and VDSR) which use interpolation as pre-scaling, lose the original structure and generate wrong patterns. Our HRAN results are more similar to EDSR, but unlike EDSR, HRAN does not produce blurry effects. Similarly, in `TaiyouNiSmash' image, we observe that most of the methods could not recover the tiny lines clearly and lose the structures and the blurry effect is also evident in most of the methods.

{\small
\begin{thebibliography}{10}\itemsep=-1pt
	
	\bibitem{bevilacqua2012low}
	M.~Bevilacqua, A.~Roumy, C.~Guillemot, and M.~Alberi{-}Morel.
	\newblock Low-complexity single-image super-resolution based on nonnegative
	neighbor embedding.
	\newblock In {\em British Machine Vision Conference (BMVC)}, pages 1--10, 2012.
	
	\bibitem{chen2018big}
	C.-F. Chen, Q.~Fan, N.~Mallinar, T.~Sercu, and R.~Feris.
	\newblock Big-little net: An efficient multi-scale feature representation for
	visual and speech recognition.
	\newblock {\em arXiv preprint arXiv:1807.03848}, 2018.
	
	\bibitem{dong2016image}
	C.~Dong, C.~C. Loy, K.~He, and X.~Tang.
	\newblock Image super-resolution using deep convolutional networks.
	\newblock {\em IEEE transactions on pattern analysis and machine intelligence},
	38(2):295--307, 2016.
	
	\bibitem{dong2016accelerating}
	C.~Dong, C.~C. Loy, and X.~Tang.
	\newblock Accelerating the super-resolution convolutional neural network.
	\newblock In {\em European conference on computer vision}, pages 391--407.
	Springer, 2016.
	
	\bibitem{dumoulin2017learned}
	V.~Dumoulin, J.~Shlens, and M.~Kudlur.
	\newblock A learned representation for artistic style.
	\newblock {\em Proc. of ICLR}, 2, 2017.
	
	\bibitem{haris2018deep}
	M.~Haris, G.~Shakhnarovich, and N.~Ukita.
	\newblock Deep back-projection networks for super-resolution.
	\newblock In {\em IEEE Conference on Computer Vision and Pattern Recognition
		(CVPR)}, 2018.
	
	\bibitem{he2016deep}
	K.~He, X.~Zhang, S.~Ren, and J.~Sun.
	\newblock Deep residual learning for image recognition.
	\newblock In {\em Proceedings of the IEEE conference on computer vision and
		pattern recognition}, pages 770--778, 2016.
	
	\bibitem{hu2018squeeze}
	J.~Hu, L.~Shen, and G.~Sun.
	\newblock Squeeze-and-excitation networks.
	\newblock In {\em Proceedings of the IEEE conference on computer vision and
		pattern recognition}, pages 7132--7141, 2018.
	
	\bibitem{huang2015single}
	J.-B. Huang, A.~Singh, and N.~Ahuja.
	\newblock Single image super-resolution from transformed self-exemplars.
	\newblock In {\em Proceedings of the IEEE Conference on Computer Vision and
		Pattern Recognition}, pages 5197--5206, 2015.
	
	\bibitem{kim2016accurate}
	J.~Kim, J.~K. Lee, and K.~M. Lee.
	\newblock Accurate image super-resolution using very deep convolutional
	networks.
	\newblock In {\em The IEEE Conference on Computer Vision and Pattern
		Recognition (CVPR Oral)}, June 2016.
	
	\bibitem{Kim_2016_DRCN}
	J.~Kim, J.~K. Lee, and K.~M. Lee.
	\newblock Deeply-recursive convolutional network for image super-resolution.
	\newblock In {\em The IEEE Conference on Computer Vision and Pattern
		Recognition (CVPR Oral)}, June 2016.
	
	\bibitem{lai2017laplacian}
	W.-S. Lai, J.-B. Huang, N.~Ahuja, and M.-H. Yang.
	\newblock Deep laplacian pyramid networks for fast and accurate
	super-resolution.
	\newblock In {\em IEEE Conferene on Computer Vision and Pattern Recognition},
	2017.
	
	\bibitem{MSLapSRN}
	W.-S. Lai, J.-B. Huang, N.~Ahuja, and M.-H. Yang.
	\newblock Fast and accurate image super-resolution with deep laplacian pyramid
	networks.
	\newblock {\em arXiv:1710.01992}, 2017.
	
	\bibitem{lanaras2015hyperspectral}
	C.~Lanaras, E.~Baltsavias, and K.~Schindler.
	\newblock Hyperspectral super-resolution by coupled spectral unmixing.
	\newblock In {\em Proceedings of the IEEE International Conference on Computer
		Vision}, pages 3586--3594, 2015.
	
	\bibitem{ledig2017photo}
	C.~Ledig, L.~Theis, F.~Husz{\'a}r, J.~Caballero, A.~Cunningham, A.~Acosta,
	A.~Aitken, A.~Tejani, J.~Totz, Z.~Wang, et~al.
	\newblock Photo-realistic single image super-resolution using a generative
	adversarial network.
	\newblock In {\em Proceedings of the IEEE conference on computer vision and
		pattern recognition}, pages 4681--4690, 2017.
	
	\bibitem{Li_2018_ECCV}
	J.~Li, F.~Fang, K.~Mei, and G.~Zhang.
	\newblock Multi-scale residual network for image super-resolution.
	\newblock In {\em The European Conference on Computer Vision (ECCV)}, September
	2018.
	
	\bibitem{lim2017enhanced}
	B.~Lim, S.~Son, H.~Kim, S.~Nah, and K.~M. Lee.
	\newblock Enhanced deep residual networks for single image super-resolution.
	\newblock In {\em The IEEE Conference on Computer Vision and Pattern
		Recognition (CVPR) Workshops}, July 2017.
	
	\bibitem{lin2017feature}
	T.-Y. Lin, P.~Doll{\'a}r, R.~Girshick, K.~He, B.~Hariharan, and S.~Belongie.
	\newblock Feature pyramid networks for object detection.
	\newblock In {\em Proceedings of the IEEE Conference on Computer Vision and
		Pattern Recognition}, pages 2117--2125, 2017.
	
	\bibitem{martin2001database}
	D.~Martin, C.~Fowlkes, D.~Tal, and J.~Malik.
	\newblock A database of human segmented natural images and its application to
	evaluating segmentation algorithms and measuring ecological statistics.
	\newblock In {\em null}, page 416. IEEE, 2001.
	
	\bibitem{matsui2017sketch}
	Y.~Matsui, K.~Ito, Y.~Aramaki, A.~Fujimoto, T.~Ogawa, T.~Yamasaki, and
	K.~Aizawa.
	\newblock Sketch-based manga retrieval using manga109 dataset.
	\newblock {\em Multimedia Tools and Applications}, 76(20):21811--21838, 2017.
	
	\bibitem{peleg2014statistical}
	T.~Peleg and M.~Elad.
	\newblock A statistical prediction model based on sparse representations for
	single image super-resolution.
	\newblock {\em IEEE transactions on image processing}, 23(6):2569--2582, 2014.
	
	\bibitem{sajjadi2017enhancenet}
	M.~S. Sajjadi, B.~Scholkopf, and M.~Hirsch.
	\newblock Enhancenet: Single image super-resolution through automated texture
	synthesis.
	\newblock In {\em Proceedings of the IEEE International Conference on Computer
		Vision}, pages 4491--4500, 2017.
	
	\bibitem{sandler2018mobilenetv2}
	M.~Sandler, A.~Howard, M.~Zhu, A.~Zhmoginov, and L.-C. Chen.
	\newblock Mobilenetv2: Inverted residuals and linear bottlenecks.
	\newblock In {\em Proceedings of the IEEE Conference on Computer Vision and
		Pattern Recognition}, pages 4510--4520, 2018.
	
	\bibitem{seferbekov2018feature}
	S.~Seferbekov, V.~Iglovikov, A.~Buslaev, and A.~Shvets.
	\newblock Feature pyramid network for multi-class land segmentation.
	\newblock In {\em The IEEE Conference on Computer Vision and Pattern
		Recognition (CVPR) Workshops}, 2018.
	
	\bibitem{shi2016real}
	W.~Shi, J.~Caballero, F.~Husz{\'a}r, J.~Totz, A.~P. Aitken, R.~Bishop,
	D.~Rueckert, and Z.~Wang.
	\newblock Real-time single image and video super-resolution using an efficient
	sub-pixel convolutional neural network.
	\newblock In {\em Proceedings of the IEEE conference on computer vision and
		pattern recognition}, pages 1874--1883, 2016.
	
	\bibitem{shi2013cardiac}
	W.~Shi, J.~Caballero, C.~Ledig, X.~Zhuang, W.~Bai, K.~Bhatia, A.~M. S.~M.
	de~Marvao, T.~Dawes, D.~O’Regan, and D.~Rueckert.
	\newblock Cardiac image super-resolution with global correspondence using
	multi-atlas patchmatch.
	\newblock In {\em International Conference on Medical Image Computing and
		Computer-Assisted Intervention}, pages 9--16. Springer, 2013.
	
	\bibitem{Tai-DRRN-2017}
	Y.~Tai, J.~Yang, and X.~Liu.
	\newblock Image super-resolution via deep recursive residual network.
	\newblock In {\em Proceedings of the IEEE Conference on Computer Vision and
		Pattern Recognition}, 2017.
	
	\bibitem{tai2017memnet}
	Y.~Tai, J.~Yang, X.~Liu, and C.~Xu.
	\newblock Memnet: A persistent memory network for image restoration.
	\newblock In {\em Proceedings of International Conference on Computer Vision},
	2017.
	
	\bibitem{tan2018feature}
	W.~Tan, B.~Yan, and B.~Bare.
	\newblock Feature super-resolution: Make machine see more clearly.
	\newblock In {\em Proceedings of the IEEE Conference on Computer Vision and
		Pattern Recognition}, pages 3994--4002, 2018.
	
	\bibitem{timofte2017ntire}
	R.~Timofte, E.~Agustsson, L.~Van~Gool, M.-H. Yang, and L.~Zhang.
	\newblock Ntire 2017 challenge on single image super-resolution: Methods and
	results.
	\newblock In {\em Proceedings of the IEEE Conference on Computer Vision and
		Pattern Recognition Workshops}, pages 114--125, 2017.
	
	\bibitem{Timofte_2016_CVPR}
	R.~Timofte, R.~Rothe, and L.~Van~Gool.
	\newblock Seven ways to improve example-based single image super resolution.
	\newblock In {\em The IEEE Conference on Computer Vision and Pattern
		Recognition (CVPR)}, June 2016.
	
	\bibitem{tong2017image}
	T.~Tong, G.~Li, X.~Liu, and Q.~Gao.
	\newblock Image super-resolution using dense skip connections.
	\newblock In {\em Proceedings of the IEEE International Conference on Computer
		Vision}, pages 4799--4807, 2017.
	
	\bibitem{wang2004image}
	Z.~Wang, A.~C. Bovik, H.~R. Sheikh, E.~P. Simoncelli, et~al.
	\newblock Image quality assessment: from error visibility to structural
	similarity.
	\newblock {\em IEEE transactions on image processing}, 13(4):600--612, 2004.
	
	\bibitem{wang2015deep}
	Z.~Wang, D.~Liu, J.~Yang, W.~Han, and T.~Huang.
	\newblock Deep networks for image super-resolution with sparse prior.
	\newblock In {\em Proceedings of the IEEE International Conference on Computer
		Vision}, pages 370--378, 2015.
	
	\bibitem{yang2010image}
	J.~Yang, J.~Wright, T.~S. Huang, and Y.~Ma.
	\newblock Image super-resolution via sparse representation.
	\newblock {\em IEEE transactions on image processing}, 19(11):2861--2873, 2010.
	
	\bibitem{yu2017dilated}
	F.~Yu, V.~Koltun, and T.~Funkhouser.
	\newblock Dilated residual networks.
	\newblock In {\em Proceedings of the IEEE conference on computer vision and
		pattern recognition}, pages 472--480, 2017.
	
	\bibitem{zeyde2010single}
	R.~Zeyde, M.~Elad, and M.~Protter.
	\newblock On single image scale-up using sparse-representations.
	\newblock In {\em International conference on curves and surfaces}, pages
	711--730. Springer, 2010.
	
	\bibitem{zhang2017learning}
	K.~Zhang, W.~Zuo, S.~Gu, and L.~Zhang.
	\newblock Learning deep cnn denoiser prior for image restoration.
	\newblock In {\em IEEE Conference on Computer Vision and Pattern Recognition},
	pages 3929--3938, 2017.
	
	\bibitem{zhang2018learning}
	K.~Zhang, W.~Zuo, and L.~Zhang.
	\newblock Learning a single convolutional super-resolution network for multiple
	degradations.
	\newblock In {\em IEEE Conference on Computer Vision and Pattern Recognition},
	pages 3262--3271, 2018.
	
	\bibitem{zhang2018image}
	Y.~Zhang, K.~Li, K.~Li, L.~Wang, B.~Zhong, and Y.~Fu.
	\newblock Image super-resolution using very deep residual channel attention
	networks.
	\newblock In {\em Proceedings of the European Conference on Computer Vision
		(ECCV)}, pages 286--301, 2018.
	
	\bibitem{zhang2018residual}
	Y.~Zhang, Y.~Tian, Y.~Kong, B.~Zhong, and Y.~Fu.
	\newblock Residual dense network for image super-resolution.
	\newblock In {\em CVPR}, 2018.
	
	\bibitem{zhang2019dcsr}
	Z.~Zhang, X.~Wang, and C.~Jung.
	\newblock Dcsr: Dilated convolutions for single image super-resolution.
	\newblock {\em IEEE Transactions on Image Processing}, 28(4):1625--1635, 2019.
	
	\bibitem{zou2012very}
	W.~W. Zou and P.~C. Yuen.
	\newblock Very low resolution face recognition problem.
	\newblock {\em IEEE Transactions on image processing}, 21(1):327--340, 2012.
	
\end{thebibliography}

}

\end{document}